\documentclass[lettersize,journal]{IEEEtran}
\usepackage{amsmath,amsfonts}
\usepackage{algorithmic}
\usepackage{algorithm}
\usepackage{array}
\usepackage[caption=false,font=normalsize,labelfont=sf,textfont=sf]{subfig}
\usepackage{textcomp}
\usepackage{stfloats}
\usepackage{url}
\usepackage{verbatim}
\usepackage{graphicx}
\usepackage{cite}
\usepackage{amsmath}
\usepackage{epstopdf}
\usepackage{graphicx}
\usepackage{amsmath}
\usepackage{amsbsy}
\usepackage{amssymb}
\usepackage{booktabs}
\usepackage{multirow}
\usepackage{arydshln}
\usepackage{indentfirst}

\usepackage{algorithm, algorithmic}
\usepackage{pifont}
\usepackage{bbding}
\usepackage{gensymb}
\usepackage{colortbl}
\usepackage{bbm}

\begin{document}

\title{ATR-UMMIR: A Benchmark Dataset for UAV-Based Multimodal Image Registration under Complex Imaging Conditions}


\author{Kangcheng~Bin, Chen~Chen, Ting~Hu, Jiahao~Qi, and Ping~Zhong, Senior Member, IEEE
\thanks{This work was supported by the National Natural Science Foundation of China under Grant 62201586. (*Corresponding author: Ping~Zhong.)

All the authors are the School of Electronic Science and Technology, National University of Defense Technology, Changsha 410003, China (email: binkc21@nudt.edu.cn, chenchen21c@nudt.edu.cn, huting@nudt.edu.cn, qijiahao1996@nudt.edu.cn, zhongping@nudt.edu.cn).

}}

\markboth{Journal of \LaTeX\ Class Files,~Vol.~14, No.~8, August~2021}%
{Shell \MakeLowercase{\textit{et al.}}: A Sample Article Using IEEEtran.cls for IEEE Journals}

\IEEEpubid{0000--0000/00\$00.00~\copyright~2021 IEEE}

\maketitle

\begin{abstract}
Multimodal fusion has become a key enabler for UAV-based object detection, as each modality provides complementary cues for robust feature extraction.
However, due to significant differences in resolution, field of view, and sensing characteristics across modalities, accurate registration is a prerequisite before fusion.
Despite its importance, there is currently no publicly available benchmark specifically designed for multimodal registration in UAV-based aerial scenarios, which severely limits the development and evaluation of advanced registration methods under real-world conditions.
To bridge this gap, we present ATR-UMMIR, the first benchmark dataset specifically tailored for multimodal image registration in UAV-based applications.
This dataset includes 7,969 triplets of raw visible (1920×1080), infrared (640×512), and precisely registered visible (640×512) images captured covers diverse scenarios including flight altitudes from 80m to 300m, camera angles from 0° to 75°, and all-day, all-year temporal variations under rich weather and illumination conditions.
To ensure high registration quality, we design a semi-automated annotation pipeline to introduce reliable pixel-level ground truth to each triplet.
In addition, each triplet is annotated with six imaging condition attributes, enabling benchmarking of registration robustness under real-world deployment settings.
To further support downstream tasks, we provide object-level annotations on all registered images, covering 11 object categories with 77,753 visible and 78,409 infrared bounding boxes.
This makes ATR-UMMIR the first dataset to simultaneously support pixel-level registration and object-level detection evaluation in UAV-based multimodal settings.
We believe ATR-UMMIR will serve as a foundational benchmark for advancing multimodal registration, fusion, and perception in real-world UAV scenarios.
The datatset can be download from \textit{https://github.com/supercpy/ATR-UMMIR}.
\end{abstract}



\section{Introduction}
Multimodal image understanding, particularly the fusion of visible and infrared imagery, is crucial for achieving robust aerial perception in complex environments \cite{DBLP:journals/inffus/TangYM22, DBLP:journals/corr/abs-2203-05406}.
UAV-based sensing introduces unique challenges: varying altitudes, wide field-of-view differences, and distinct imaging resolutions between visible (e.g., 1920×1080) and infrared (e.g., 640×512) sensors \cite{DBLP:journals/tcsv/SunCZH22}. These discrepancies hinder direct fusion and joint utilization in downstream tasks like object detection and semantic segmentation \cite{DBLP:journals/tiv/SongXWJYM24,chen2024weakly}.

However, current practice in UAV scenarios heavily depends on manual alignment, which often requires expert intervention, significant image cropping, and geometric warping. This process is labor-intensive, hard to scale, and prone to inconsistencies across datasets \cite{DBLP:journals/pami/YingXALHLCLWHXLLZLS25}.
Despite the importance of this problem, there is still a lack of systematic research in the field. The obstacle is the difficulty of collecting large-scale aerial multimodal data under comprehensive imaging conditions. Moreover, acquiring accurate pixel-level ground-truth correspondences between visible and infrared images in aerial settings is technically challenging and labor-intensive. As a result, no publicly available dataset currently supports multimodal registration for UAV-based imaging, limiting progress on this fundamental yet underexplored problem.

To advance multimodal image registration under real-world UAV imaging conditions, we introduce ATR-UMMIR, the first publicly available benchmark dataset focusing on UAV-based multimodal image registration. 
It comprises 7,969 triplets of registered image samples, where each triplet includes one high-resolution raw visible image (1920×1080), one raw infrared image (640×512), and one registered visible image (640×512) precisely registered to the infrared view. 
To gain high-quality registered visible image for each raw multimodal image pairs, we develop a semi-automated registration pipeline that involves keyframe selection, manual temporal synchronization, coarse spatial alignment, and fine-grained automatic registration, yielding accurate pixel-level ground truth.
Our ATR-UMMIR offers the following key contributions:

\textbf{(1) Cross-Resolution and FOV Registration Benchmarking:}
ATR-UMMIR is the first dataset that explicitly addresses the challenge of multimodal registration across different resolutions and fields of view. 
It supports the evaluation of both rigid and non-rigid registration methods under resolution and FOV variation.

\textbf{(2) Comprehensive Imaging Conditions with Attribute Annotations:}
The dataset covers diverse aerial imaging conditions, including flight altitudes from 80m to 300m, camera angles from 0° to 75°, and various times of day, seasons, weather, and illumination states. Each image pair is annotated with six key condition attributes, enabling condition-aware evaluation and reflecting the data distribution of real-world UAV deployments.

\textbf{(3) Downstream Utility in Multimodal Perception:}
To validate the utility of registration in downstream tasks, ATR-UMMIR provides fine-grained object bounding box annotations: 77,753 for visible and 78,409 for infrared images. This allows direct assessment of how registration quality affects multimodal detection and fusion-based recognition tasks.

\begin{figure*}[htb!]
  \centering
   \includegraphics[width=0.99\linewidth]{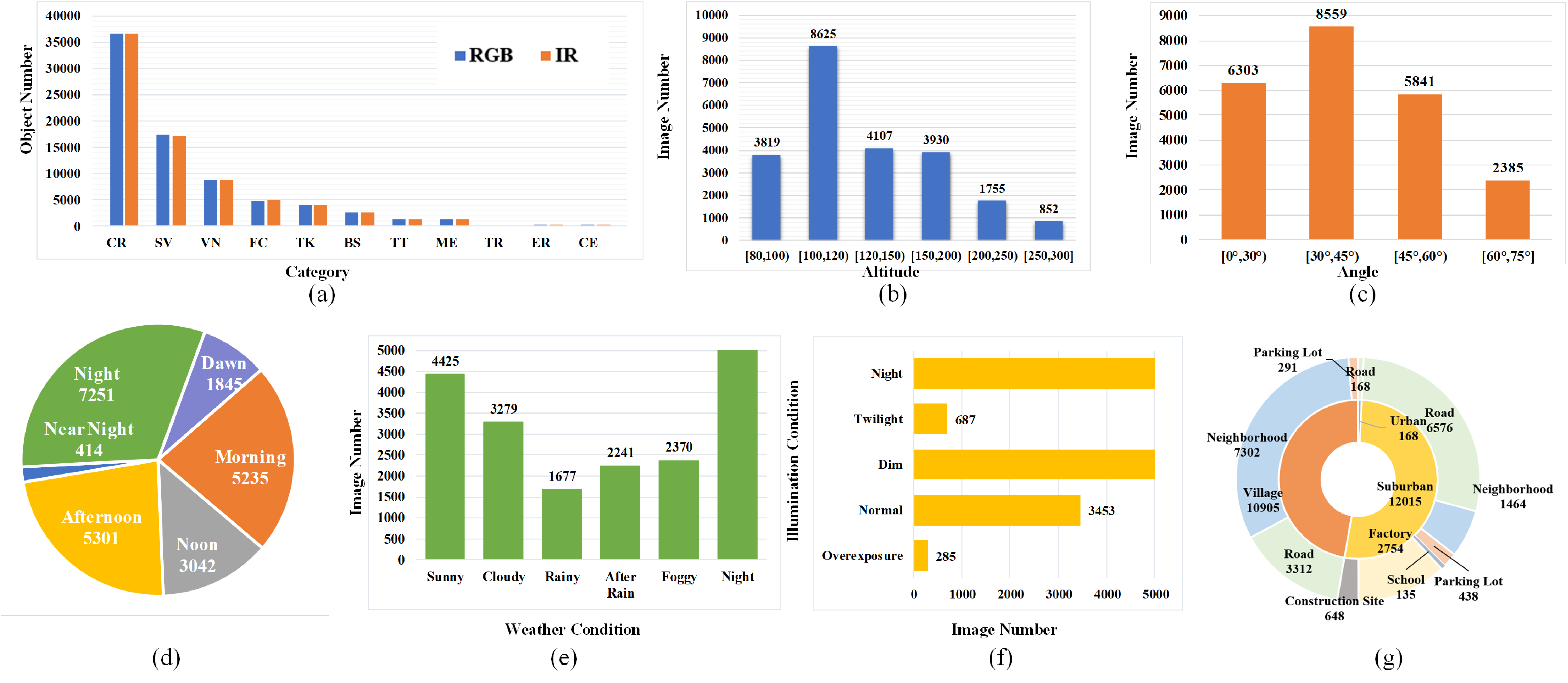}
   \vspace{-0.4cm}
   \caption{The object and attribute statistics of the ATR-UMMIR dataset. Note that CR, SV, VN, BS, FC, TK, ME, TR, ER, CE  and TT represent car, SUV, van, bus, freight car, truck, motorcycle, trailer, excavator, crane, and tank truck categories, respectively.}
   \label{fig:1} 
   \vspace{-0.6cm}
\end{figure*}
\label{sec:3}

\section{ATR-UMMIR Dataset}
\subsection{Dataset Construction}
\textbf{Data collection.}
The ATR-UMMIR dataset encompasses variations in flying altitude, camera angle, shooting time, weather, illumination, and scenario. 
The data were collected using synchronized visible and infrared cameras mounted on DJI H20T and DJI H20N UAV platforms. To obtain accurate ground-truth registrations, we developed a semi-automatic registration framework. This process includes keyframe selection, manual temporal synchronization, expert-guided coarse spatial warping, and fine-grained automatic refinement.

\textbf{Object annotation.}
For annotation, objects in RGB and IR modalities were independently labeled using oriented bounding boxes to precisely capture their positions and orientations. This modality-specific annotation strategy ensures accurate object localization across varying spectral characteristics and facilitates fine-grained evaluation of cross-modal registration and detection performance.

\textbf{Attribute annotation.}
To support a deeper understanding of how environmental factors affect multimodal image matching and fusion, we augment the ATR-UMMIR dataset with comprehensive condition annotations. 
Each image pair is annotated with six critical imaging attributes—\textit{Altitude}, \textit{Angle}, \textit{Time}, \textit{Weather}, \textit{Illumination}, and \textit{Scenario}. These condition labels provide vital contextual information, enabling researchers to investigate performance bottlenecks under specific conditions and to develop condition-aware fusion strategies for robust perception.

\subsection{Dataset Statistics}

\textbf{Image statistics.}
The ATR-UMMIR dataset consists of 7,969 triplets of UAV-captured images. Each triplet contains one high-resolution raw visible image (1920×1080), one raw infrared image (640×512), and one registered visible image (640×512) that has been spatially aligned with the infrared counterpart. This triplet structure enables both unimodal and multimodal tasks, supporting research in image registration, cross-modal matching, and fusion-based recognition.

\textbf{Object statistics.}
ATR-UMMIR provides fine-grained object annotations in the form of oriented bounding boxes, with a total of 77,753 instances annotated in visible images and 78,409 in infrared images. These annotations were performed independently on each modality to account for modality-specific appearance variations. The availability of dense and modality-aware annotations enables direct evaluation of how registration accuracy influences the performance of cross-modal object detection and fusion-based recognition. Furthermore, it supports robust benchmarking of detection consistency, alignment robustness, and feature-level fusion strategies under varying conditions.

\textbf{Attribution statistics.}
To comprehensively reflect real-world variability, the ATR-UMMIR dataset offers condition annotations across six key imaging attributes. As shown in Fig.~\ref{fig:1}, flight altitudes range from 80 to 300 meters, with most images captured at 100–120m (8,625), ensuring realistic aerial perspectives. Camera angles span from nadir to 75°, with a dominant portion between 30°–45° (8,559), providing rich viewpoint diversity. The dataset covers a complete diurnal cycle, including 7,251 nighttime images and thousands captured in morning, afternoon, and dawn, enabling temporal robustness analysis. Weather conditions include sunny, cloudy, rainy, after-rain, foggy, and night scenes, representing typical and adverse outdoor scenarios. Illumination levels are labeled into night, twilight, dim, normal, and overexposure, with over 4,000 low-light or night images supporting evaluation of modality complementarity under challenging lighting. Scenario-wise, the dataset spans 11 categories such as urban, suburban, village, road, neighborhood, factory, and parking lot, ensuring environmental complexity and supporting robust, condition-aware multimodal registration research.

\section{Conclusion}
In this work, we present ATR-UMMIR, the first benchmark dataset dedicated to UAV-based multimodal image registration. Addressing a long-standing gap in the field, ATR-UMMIR offers 7,969 triplets of registered visible and infrared images captured under diverse and challenging aerial conditions. Through a carefully designed semi-automated registration pipeline, we provide high-quality pixel-level ground truth despite differences in resolution and field of view between modalities.


\bibliographystyle{IEEEtran}
\bibliography{ref.bib}

\end{document}